\documentclass[runningheads]{llncs}
\usepackage{graphicx}
\usepackage{tabu}
\usepackage{bbm}
\usepackage{amsmath}
\usepackage{caption}
\usepackage{subcaption}
\usepackage{color}
\usepackage{colortbl}
\usepackage{xcolor}
\usepackage{hyperref}

\begin{document}
\title{MAD: Modality Agnostic Distance Measure for Image Registration}
\titlerunning{MAD: Modality Agnostic Distance Measure for Image Registration}
%
\author{
Vasiliki {Sideri-Lampretsa}\inst{1, 2} \and Veronika A. Zimmer\inst{1} \and Huaqi Qiu\inst{4} \and Georgios Kaissis\inst{1, 2, 3} \and Daniel Rueckert \inst{1, 2, 4}}

\authorrunning{Vasiliki Sideri-Lampretsa et al.}
\institute{Technical University of Munich \\
\email{\{vasiliki.sideri-lampretsa, veronika.zimmer, g.kaissis, daniel.rueckert\}@tum.de} \\
\and Klinkum rechts der Isar, Munich, Germany \and
Helmholtz Zentrum Munich, Germany \and Department of Computing, Imperial College London, London, UK 
\email{huaqi.qiu15@imperial.ac.uk}
}
%
\maketitle              
\begin{abstract}
Multi-modal image registration is a crucial pre-processing step in many medical applications. However, it is a challenging task due to the complex intensity relationships between different imaging modalities, which can result in large discrepancy in image appearance. The success of multi-modal image registration, whether it is conventional or learning based, is predicated upon the choice of an appropriate distance (or similarity) measure. Particularly, deep learning registration algorithms lack in accuracy or even fail completely when attempting to register data from an "unseen" modality. In this work, we present Modality Agnostic Distance (MAD), a deep \textit{image distance} measure that utilises random convolutions to learn the inherent geometry of the images while being robust to large appearance changes. Random convolutions are geometry-preserving modules which we use to simulate an infinite number of synthetic modalities alleviating the need for aligned paired data during training. We can therefore train MAD on a mono-modal dataset and successfully apply it to a multi-modal dataset. We demonstrate that not only can MAD affinely register multi-modal images successfully, but it has also a larger capture range than traditional measures such as Mutual Information and Normalised Gradient Fields. Our code is available at: \href{https://github.com/ModalityAgnosticDistance/MAD}{https://github.com/ModalityAgnosticDistance/MAD}.

\keywords{Image registration, mutli-modality, distance measure}
\end{abstract}

\section{Introduction}

Multi-modal image registration is a crucial and challenging application in medical image processing. It generally refers to the process in which two images acquired from different imaging systems, governed by different physics principles, are aligned into the same coordinate space. Fusing the different modalities can result in more informative content. However, this is not a trivial problem due to the highly non-linear relationships between the shapes and the appearance present in different modalities \cite{Sotiras2013DeformableMI}.

In order to tackle this challenging issue, several multi-modal image similarity/distance measures have been proposed \cite{Haber2006IntensityGB,Heinrich2012MINDMI,Studholme1999AnOI}. The widely used multi-modal intensity-based measures, Mutual Information (MI) \cite{Loeckx2010NonrigidIR,Studholme1999AnOI,Wells1996MultimodalVR}, operates on intensity histograms and is therefore agnostic to the underlying geometry of the image structures. Although MI excels in aligning images that are relatively close in space, it shows limited ability to recover large misalignments without a multi-resolution framework. Other metrics such as Normalised Gradient Fields (NGF), measure the image similarity using edge maps \cite{Haber2006IntensityGB,Wachinger2012EntropyAL} while the Modality Independent Neighborhood Descriptor (MIND) measures the image similarity using hand-crafted local descriptors \cite{Heinrich2012MINDMI,Woo2015MultimodalRV}. However, these measures make only restrictive assumptions on the intensity relationships between multi-modal images which affect their performance. Apart from the hand-crafted measures, many learning-based distance measures have also been proposed \cite{Cheng2015DeepSL,Czolbe2020DeepSimSS,Haskins2018LearningDS,Lee2009LearningSM,Simonovsky2016ADM,Wachinger2012EntropyAL}. Most of these, however, are either used only for mono-modal registration only or require ground truth transformation or pre-aligned paired multi-modal images for training which is very challenging to obtain in a real-world scenario. In this work, we propose Modality Agnostic Distance (MAD), a self-supervised contrast-agnostic geometry-informed deep distance measure that demonstrates a wider capture range than the traditional measures without using a multi-resolution scheme. We overcome the limited assumptions of intensity relations in the intensity-based distance measures by learning geometry-centric relationships with a neural network. This is achieved by using random convolutions to create complex appearance changes, which also enables us to synthesise infinite aligned image pairs of different modalities, alleviating the need for aligned multi-modal paired data in the existing learning-based distance measures. To the best of our knowledge, our work is the first that explores random convolutions as data augmentation in the context of medical imaging and medical image registration. Our contribution can be summarised as follows:
\begin{itemize}
\item[$\bullet$] We introduce learning a general geometry-aware contrast-agnostic deep \textit{image distance} measure for multi-modal registration, formulating an effective self-supervised task that allows the network to assess the \textit{image distance} by grasping the underlying geometry independent of its appearance. 

\item[$\bullet$] We propose to use random convolutions to obtain infinite aligned image pairs of different appearances and parametric geometry augmentation to learn a modality-invariant distance measure for image registration. 

\item[$\bullet$] We perform a detailed study about the capture range and evaluate the effectiveness of the proposed measure through extensive ablation analysis on two challenging multi-modal registration tasks, namely multi-modal brain registration and Computed Tomography (CT) - Magnetic Resonance (MR) abdominal image registration.

\end{itemize}

\begin{figure}
\centering
\includegraphics[width=\textwidth]{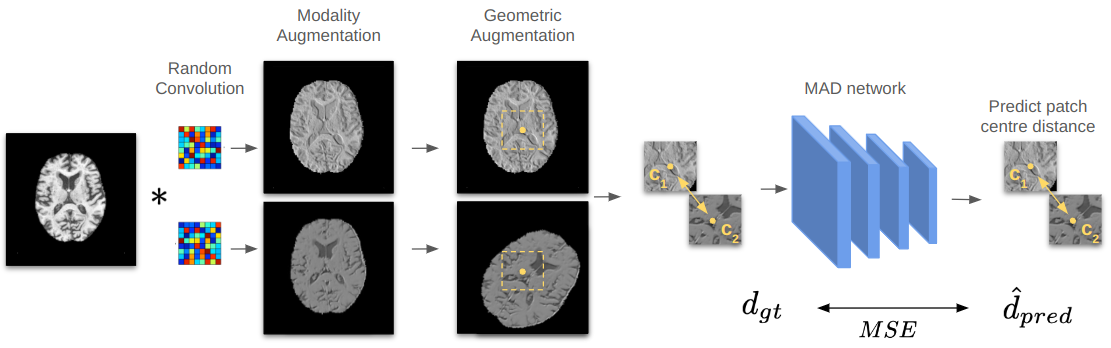}
\caption{An overview of our method. We synthesise modalities from a mono-modal dataset using random convolutions (left). Random affine transformations are used as geometric augmentation (middle). Patches are sampled at corresponding locations as input to a CNN to regress the distance between patch centres (right). We compute the MSE between ground truth (\(d_{gt}\)) and predicted (\(d_{gt}\)) distance.} \label{fig1}
\end{figure}

\section{Related Works}


Besides the hand-crafted measures described in the introduction, learning-based methods were also proposed to learn an inter-modal loss function \cite{Cheng2015DeepSL,Czolbe2020DeepSimSS,Haskins2018LearningDS,Lee2009LearningSM,Simonovsky2016ADM,Wachinger2012EntropyAL}. A recent work, DeepSim~\cite{Czolbe2020DeepSimSS} proposes to pre-train a feature extractor on an auxiliary segmentation task and then use the semantic features to drive the optimisation of a learning-based registration model. Moreover, Pielawski et al. \cite{Pielawski2020CoMIRCM} proposed CoMIR that uses supervised contrastive learning and paired data to learn affine registration on 2D images. Dey et al. \cite{Dey2022ContraRegCL} proposed a method that involves unsupervised contrastive learning to maximise the mutual information between features. Hoffman et al. \cite{Hoffmann2020SynthMorphLC} proposed a data augmentation pipeline based on labels to simulate synthetic images and assess the image similarity in them. Similar to our work, \cite{Simonovsky2016ADM} proposes to learn a metric from patches through a patch-matching classification task. However, the training process relies on aligned paired data which is difficult to acquire for any modality and the model is specifically trained on T1w-T2w images, limiting its generalisability across different domains such as MR-CT.

\section{Methods}\label{sec:Methods}
\textbf{Problem formulation.}\label{sec:registration}
In this paper, we are focusing on affine registration between 3D images. Affine image registration is the task of estimating the best transformation \(\mathcal{A}:\Omega_{F} \rightarrow \Omega_{M}\),  between a fixed \(F: \Omega_F\subseteq \mathbbm{R}^n \rightarrow \mathbbm{R}\) and a moving image \(M:\Omega_M\subseteq \mathbbm{R}\rightarrow \mathbbm{R}^n\) ($n=3$ in our case).
In conventional registration, the transformation parameters $\mu$ that parameterise the affine matrix, here denoted by \(\mathcal{A}_{\mu}\), are estimated by solving the following optimisation problem:
\begin{equation} \label{eq:1}
  \mu^*= \arg \min_{\mu} [\mathcal{D}(F, M(\mathcal{A}_{\mu}))],
\end{equation}
\noindent where \(\mathcal{D}\) is a distance measure which measures how well the images are aligned and \(M(\mathcal{A}_{\mu})\) is the affinely transformed moving image resampled to \(\Omega_{F}\).

In learning-based registration, this problem is solved by optimising the weights of a neural network instead of the parameters of the transformation:
\begin{equation} \label{eq:2}
  \phi^*= \arg \min_{\phi}  \mathbbm{E}[\mathcal{D}(F, M(\mathcal{A}_{{g}_{\phi}}))],
\end{equation}
\noindent where \(g_{\phi}(F,M)\) is a neural network that predicts the parameters of the affine transformation which aligns the images $F$ and $M$. In both conventional and learning-based approaches, selecting an appropriate measure \(\mathcal{D}\) is crucial.

\noindent\textbf{Learning modality-agnostic distance measure.}
 Instead of using an analytically formulated distance measure, we propose to formulate $\mathcal{D}$ as a geometry-aware convolutional neural network that estimates the dissimilarity between images by aggregating the distance between their patches, while remaining contrast agnostic. In other words, we are training the network to estimate the distance between the patch centres sampled from the different images after augmenting using random convolutions. Intuitively, the centre point difference can serve as a similarity indicator, i.e. if the points are close in space that means that the patches should also be close in space and vice versa. The whole process can be schematically outlined in Fig.~\ref{fig1}.

\noindent\textbf{Modality augmentation.}\label{sec:RandConv} In order to reliably achieve multi-modal registration, we would like to devise a dissimilarity measure which is modality-agnostic, removing the need for retraining for every modality pairing. To achieve that, we propose to employ randomly initialised convolutional layers as data augmentation~\cite{Xu2021RobustAG}. These layers have the desirable trait that they maintain the underlying geometry of the images, only transforming the local intensity and texture.


Our first task is to extend the formulation presented in \cite{Xu2021RobustAG} to 3D. Convolution is a linear operator, as a result the intensities are linearly mapped to the range \([0, 255]\). This is rather unrealistic as we are often dealing with modality pairs with non-linear intensity relationships. Therefore, we introduce non-linearity in the intensity mapping by performing clamping, taking the absolute value of the result of the random convolutions and passing it through a leaky ReLU with a variable negative slope. This simulates more sophisticated intensity relationships between the augmented domains.
The geometric-preserving augmentation enforces identical structure, but different appearance enabling us to generate infinite pairs of aligned modalities only from one mono-modal image, removing the need for paired and pre-aligned multi-modal data.


\noindent\textbf{Geometric augmentation.}\label{sec:DataAug} Let $F = \text{RandConv}(I)$, $M = \text{RandConv}(I)$ be the fixed and moving image volumes which are the same image mapped to different augmented modalities via random convolutions. To train the learned distance metric, we synthetically transform the moving image to generate controlled geometric misalignment. Following the notation suggested by \cite{Mok2022AffineMI}, we sample a random rotation, translation, scaling and shearing parameter from a range of possible configurations and we construct an affine matrix $\mathcal{A}$. Formally, we can write the affine matrix \(\mathcal{A}(\mathbf{t}, \mathbf{r}, \mathbf{s}, \mathbf{h})\) where \(\mathbf{t}, \mathbf{r}, \mathbf{s}, \mathbf{h} \in \mathbbm{R}^3\) are the translation, rotation, scaling and shearing parameters. The affine matrix $\mathcal{A}$ can be composed by a set of geometric transformation matrices: \(A = \mathcal{T} \cdot \mathcal{R} \cdot \mathcal{S} \cdot \mathcal{H}\), where $\mathcal{T}$, $\mathcal{R}$, $\mathcal{S}$ and $\mathcal{H}$ denote the translation, rotation, scaling and shearing transformation matrices parameterised by the corresponding geometric transformation parameters.

Finally, we sample $N$ patches of the same size at the same locations in the fixed and transformed moving image resulting in $N$ patch pairs that differ both in appearance and geometry. We are denoting the patches that are sampled from the fixed image  by \(P_{F}\) and the patches that are sampled from the moving image by \(P_{M}\). Since we synthetically transform the moving image with a known affine transformation relative to the fixed, we can also obtain the ground truth deformation field which effectively denote the distance between the patch centres.

\noindent\textbf{MAD.}\label{sec:MAD} 
To construct an alignment distance measure from the patches, we employ a convolutional neural network (ResNet~\cite{he2016deep}) \(f_{\theta}\) which is trained to determine the Euclidean distance between the centres of a patch pair from their appearances. The distance measure between the images can be thus calculated by aggregating the distances between all patches: 
\begin{equation}\label{eq:3}
\mathcal{D}_{MAD}(F, M) = \frac{1}{N}\sum_{i=1}^{N}f_{\theta}(P_{F}^{i}, P_{M}^{i}),
\end{equation}
\noindent where $\theta$ are the network parameters, $N$ is the number of patches, \(P_F^i:\Omega^P_F\subset\Omega_F \subseteq \mathbbm{R}^3\rightarrow \mathbbm{R}^3\) and \(P_M^i:\Omega^P_M\subset\Omega_M \subseteq \mathbbm{R}^3\rightarrow \mathbbm{R}^3\) are patches of the same size sampled at the same location from the fixed and moving image respectively.








We are supervising the training of the patch distance CNN using the patches we generated from the modality and geometric augmentations. Concretely, we optimise the network using a loss function which calculates the Mean Square Error (MSE) between the ground truth and predicted centre point distances:
\begin{equation}
    \theta^*= \arg \min_{\theta} \mathbbm{E} [(d_{gt} - \hat{d}_{pred})^2]
\end{equation}
\noindent where $\theta^*$ denotes the optimal network parameters, \(\hat{d}_{pred}\) denotes the patch distances predicted by the network and \(d_{gt}\) denotes the ground truth distances derived from the known affine transformation \(\mathcal{A}(\mathbf{t}, \mathbf{r}, \mathbf{s}, \mathbf{h})\) that we sampled.



Intuitively, by presenting the network with a large number of augmented modality pairs with varying intensity relationships, we encourage it to focus on evaluating the difference between image shapes according to their inherent geometric transformations, while placing less emphasis on image appearance. 
Given that our distance measure is differentiable by construction, it can be optimised using gradient-based optimisation techniques and used as a cost function in both conventional or learning-based registration algorithms.


\section{Experiments \& Discussion}




\noindent\textbf{Datasets.} We are evaluating the effectiveness of the proposed measure using three datasets: the Cambridge Centre for Ageing and Neuroscience project (CamCAN) \cite{Shafto2014TheCC,Taylor2017TheCC}, the Retrospective Image Registration Evaluation Project (R.I.R.E.) \cite{West1996ComparisonAE} and the arguably challenging MR-CT dataset of Learn2Reg\cite{Hering2021Learn2RegCM}. 
We normalise all the brain images to a common MNI space using affine registration, ensuring an isotropic spatial resolution with a voxel size of \(1\rm{mm}^3\). We perform skull-stripping using ROBEX \cite{Iglesias2011Robex} and bias-field correction using the N4 algorithm in SimpleITK \cite{Lowekamp2013TheDO}. For the pre-processing of the CT images in R.I.R.E, we use the steps proposed in \cite{Muschelli2015ValidatedAB}. We utilise the $310$ 3D T1w brain images of the CamCAN project to train MAD (80\% training - 20\% validation) and we test it on 6 subjects of R.I.R.E. that have uncorrupted T1w, T2w, PD MR and CT brain images.
Regarding the abdominal images, we use the 90 unpaired, affinely aligned MR and CT scans for training (80\% training - 20\% validation) and the 16 paired scans for evaluation. All the images have isotropic spatial resolution with a voxel size of \(2\rm{mm}^3\).

\noindent\textbf{Evaluation.}
We compare MAD to two widely used image similarity measures: NMI introduced by \cite{Studholme1999AnOI}, NGF \cite{Haber2006IntensityGB} and ClassLoss (CL) which is a learning-based measure based on patch classification \cite{Simonovsky2016ADM} and most relevant to our work. We start with images from different modalities that are affinely aligned in space. Then, we transform them with a synthetic affine transformation. The intuition behind this is that by controlling the applied transformation parameters we can evaluate the measures' performance quantitatively. I.e. as we know the synthetic transformation that we are trying to recover, we can also know the ground truth deformation field. As a result, we can evaluate the accuracy of the registration by calculating the Mean Average Error (MAE) and its standard deviation (std) in \rm{mm} between the ground truth deformations and the predicted ones. We also test if the differences in the reported errors between competing methods are statistically significant (\( p < 0.05\)) with a t-test \cite{Efron2016ComputerAS}. Lastly, we calculate the effect size $es$ considering it small when \(es \leq 0.3\), moderate when \(0.3 \leq es \leq 0.5\) and strong when \(es \leq 0.5\) \cite{Cohen1969StatisticalPA}. 
We incorporate the baselines and the proposed measure with image registration implemented with Airlab, a conventional registration framework introduced in \cite{Sandkhler2018AirLabAI}. 





\subsection{Experiment 1: Loss Landscapes}\label{sec:loss_landscape}

\noindent\textbf{Setup.} We generate the measure landscapes to inspect and compare the convexity and the capture range by translating a CT image relative to a T1w image from the R.I.R.E. dataset. The translations \(\mathbf{t} = [t_x, t_y, t_z]\) are in the range of \([-60, 60]\,\rm{mm}\) with a step size of 10\rm{mm} and the resulting image distances are normalised for better comparison.
\begin{figure}
    \centering
        \begin{subfigure}[b]{0.16\textwidth}
        \includegraphics[width=\textwidth]{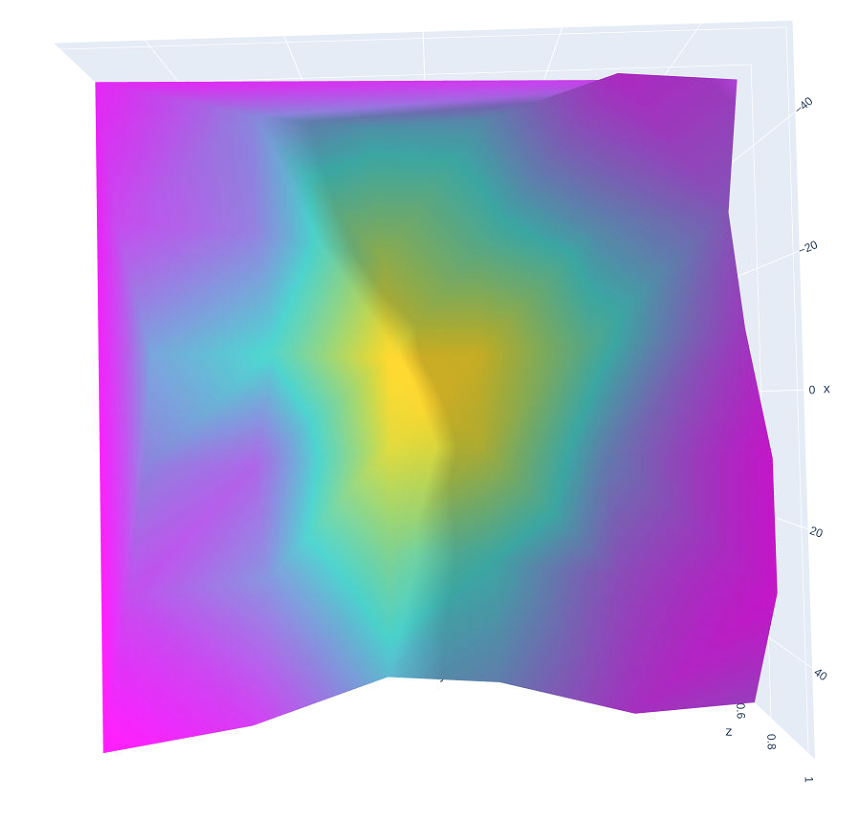}
        \caption{MAD}
        \end{subfigure}
        \begin{subfigure}[b]{0.16\textwidth}
        \includegraphics[width=\textwidth]{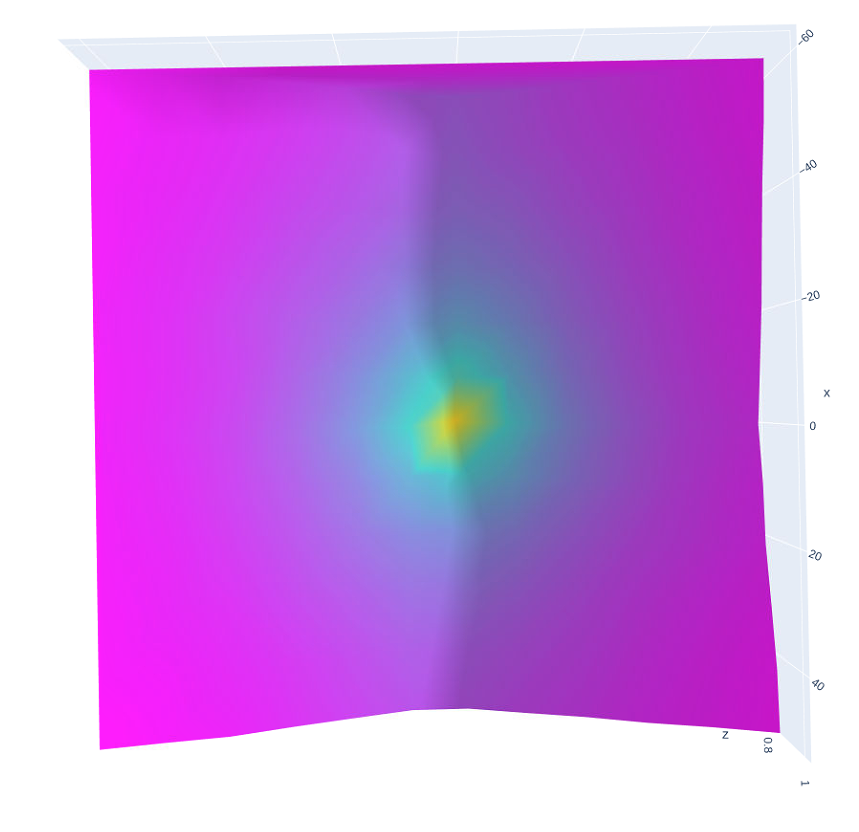}
        \caption{NGF}
        \end{subfigure}
        \begin{subfigure}[b]{0.16\textwidth}
        \includegraphics[width=\textwidth]{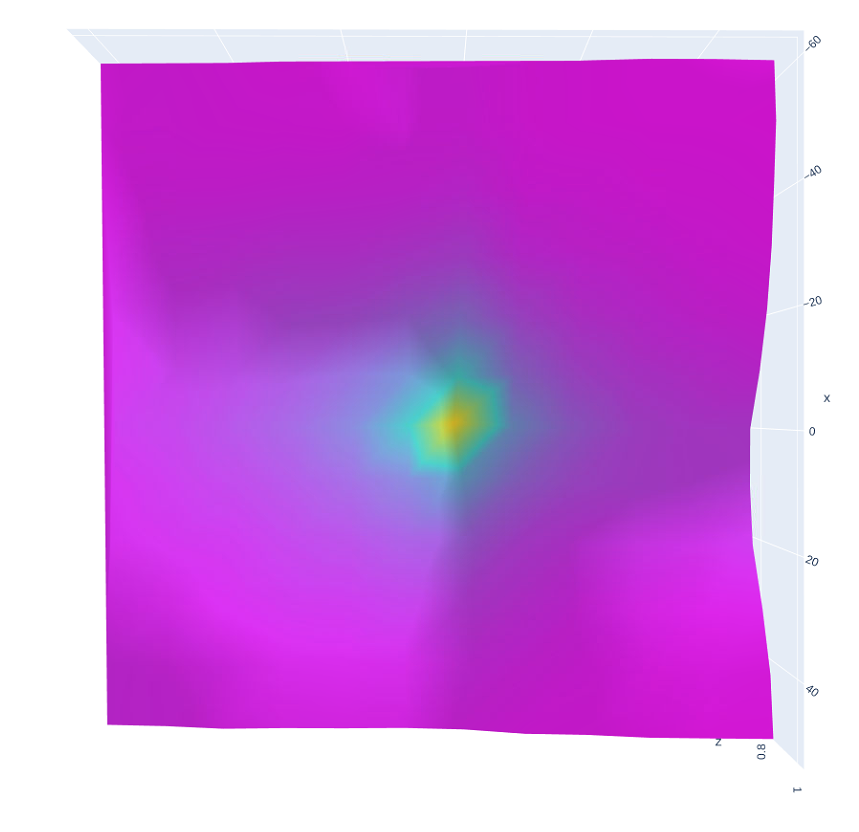}
        \caption{NMI}
        \end{subfigure}
        \begin{subfigure}[b]{0.16\textwidth}
        \includegraphics[width=\textwidth]{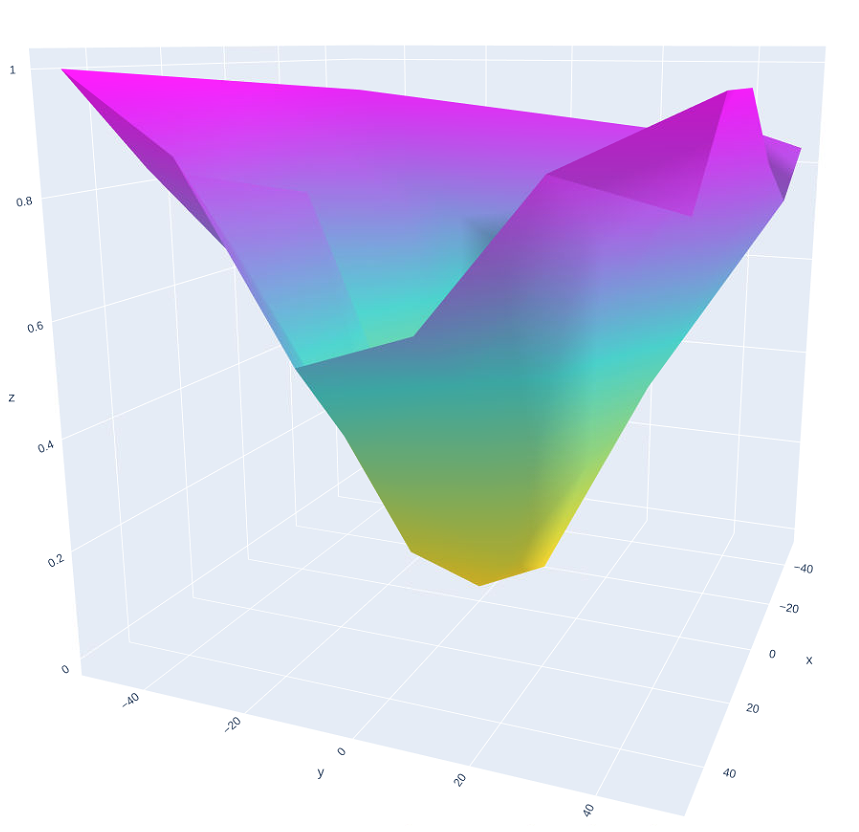}
        \caption{MAD}
        \end{subfigure}
        \begin{subfigure}[b]{0.16\textwidth}
        \includegraphics[width=\textwidth]{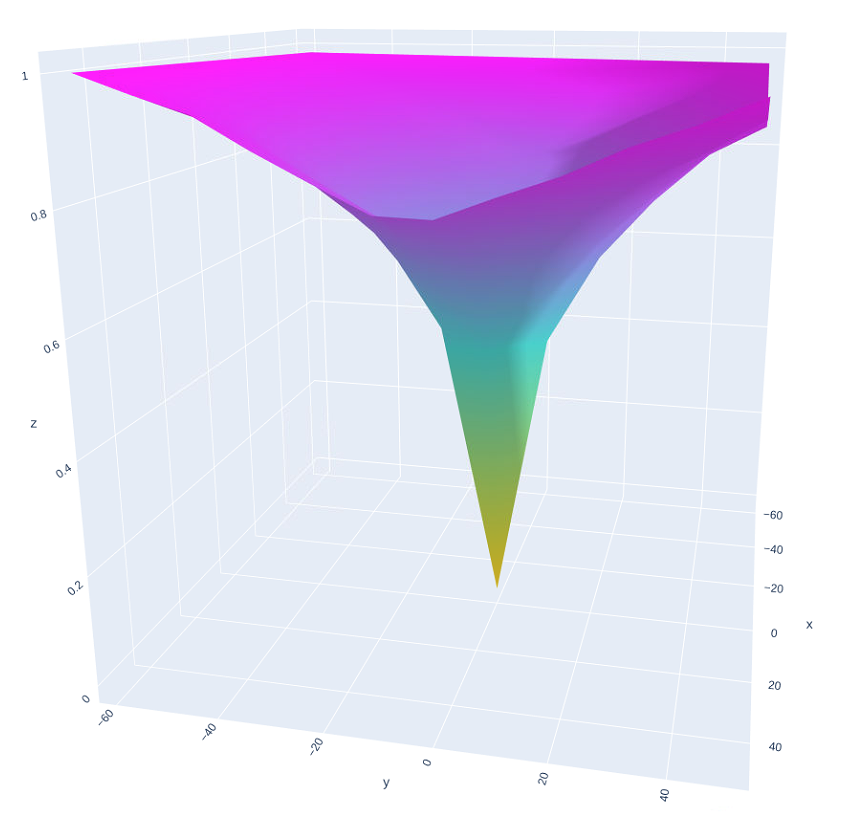}
        \caption{NGF}
        \end{subfigure}
        \begin{subfigure}[b]{0.16\textwidth}
        \includegraphics[width=\textwidth]{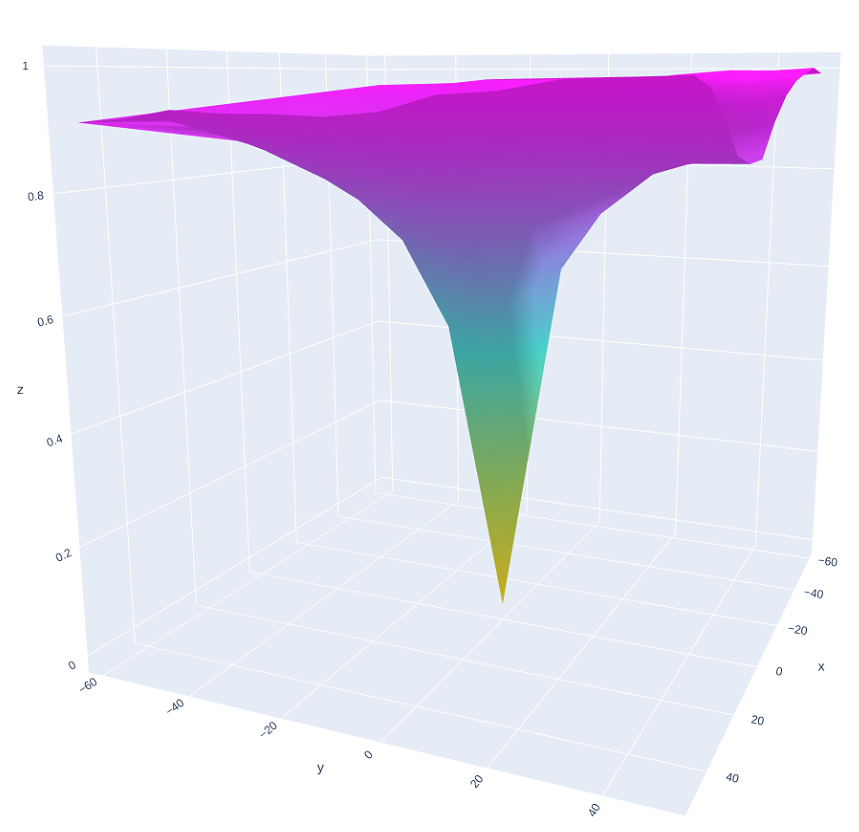}
        \caption{NMI}
        \end{subfigure}
    \caption{The loss 2D landscapes for translation in the range of \([-60, 60] \rm{mm}\). MAD (MAD) [a, d], Normalised Gradient Fields (NGF) [b, e] and  Normalised Mutual Information (NMI) [c, f]}\label{fig2}
\end{figure} 

\noindent\textbf{Results.} Figure~\ref{fig2} demonstrates that despite NMI, NGF and MAD landscapes being smooth with minima at 0, MAD leads to the largest capture range (further validated by registration results in Table \ref{tab:results}) compared to NGF and NMI. This could be explained by the fact that MAD network is trained on complex appearance patches and therefore it is able to capture the underlying geometry better than NGF that is operating on edge maps. NMI is known to perform poorly for large misalignment without the usage of a multi-resolution scheme, showing weak gradients towards the optimal alignment when the translations are large.
\subsection{Experiment 2: Recover random transformations}
\noindent\textbf{Setup.} We assess the performance of our \textit{image distance} measure in recovering synthetic affine transformations. We repeat the experiment $100$ times for all test subjects and we compute the MAE between the ground truth and the predicted deformation fields. We evaluated the performance on a small and a large range of transformations (Table \ref{tab:results} right) in order to examine the capacity of MAD to restore synthetic transformations without requiring a multi-resolution approach. For the large transformations, we employ a multi-resolution scheme for both NGF and NMI, given the small capture range these measures exhibit in Sec.~\ref{sec:loss_landscape}.
\begin{figure}[tbp]
\centering
\includegraphics[width=\textwidth]{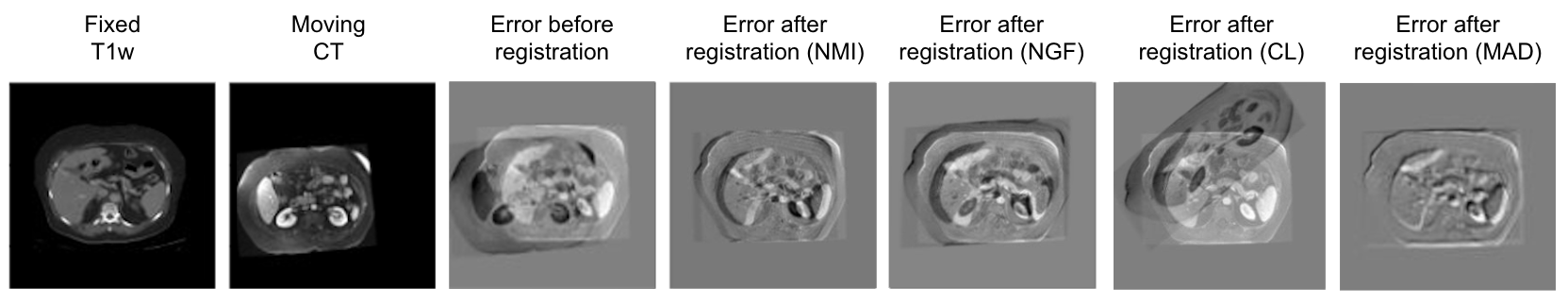}
\caption{Qualitative registration results for the MR-CT abdominal dataset using the different measures.} \label{fig:qualitative_comp}
\end{figure} 
\begin{table}[btp]
    \caption{Registration performance measured using the Mean Absolute Error (mean(std)) in \rm{mm} between the ground truth and the predicted deformation fields for small and large affine misalignment (left). Grey boxes indicate significance compared to MAD with a \(p < 0.05\) and effect sizes small, moderate ($\ast$) and strong ($\ast\ast$). Affine parameter ranges: Translation (T.), Rotation (R.), Scaling (Sc.), Shearing (Sh.) (right).}
    \begin{subtable}[h]{0.35\textwidth}
    
       \begin{tabular}{l c c c c}
        \hline
        \multicolumn{5}{c}{Small misalignment setting} \\
        \hline
        {} & T1w-PD & T1w-CT & T1w-T2w & CT-MR\\
        \hline
        NMI & \(4.79(3.22)\) & \(\mathbf{4.15(4.32)}\)  & \(\mathbf{4.07(3.51)}\)  & \(7.65(6.28)\)\\
        NGF &  $4.92(3.81)$ & $5.38(5.26)$  & $4.89(3.76)$  & \(8.81(4.29)\)\\
        CL &\(5.32(5.11)\) &\(5.46(4.38)\) & \(4.75(4.71)\)  & \(9.63(9.51)\)\\
        MAD & \(\mathbf{4.56(3.8)}\) & \(4.27(4.17)\) & \(4.15(4.12)\)  & \(\mathbf{7.06(2.07)}\)\\
        \hline
        \multicolumn{5}{c}{Large misalignment setting} \\
        \hline
        {} & T1w-PD & T1w-CT & T1w-T2w & CT-MR\\
        \hline
        NMI & \(17.79(14.41)\) & \cellcolor{gray!25} \(20.21(12.32)^{*}\)  & \(15.63(15.78)\) &\cellcolor{gray!25} \(22.93(19.72)^{*}\)\\
        NGF &\cellcolor{gray!25}$23.86(12.74)^{*}$ & \cellcolor{gray!25} $29.98(22.64)^{**}$  &\cellcolor{gray!25}  $27.88(25.15)^{*}$ & \cellcolor{gray!25} \(25.87(21.56)^{*}\)\\
        CL & \cellcolor{gray!25} \(63.24(48.28)^{**}\) & \cellcolor{gray!25} \(67.13(41.79)^{**}\) & \cellcolor{gray!25} \(55.84(49.36)^{**}\) & \cellcolor{gray!25} \(94.12(83.52)^{**}\)\\
        MAD & \(\mathbf{13.34(7.59)}\) & \(\mathbf{17.09(9.03)}\) & \(\mathbf{13.21(7.64)}\) & \(\mathbf{20.47(15.63})\)\\
        \hline
        \end{tabular}
       \label{tab:results}
    \end{subtable}
    \hfill
    \begin{subtable}[h]{0.2\textwidth}
        \centering
        \begin{tabular}{l c }
            \hline
            \multicolumn{2}{c}{Parameter ranges} \\
            \hline
            \multicolumn{2}{c}{Small} \\
            \hline
            T. & \([-30, 30]\)\rm{mm}  \\
            R. & \([-25, 25]\)\(^{\circ}\) \\
            Sc. & \([90, 110]\)\%  \\
            Sh. & \([90, 110]\)\%  \\
            \hline
            {} & {} \\
            \multicolumn{2}{c}{Large} \\
            \hline
            T. & \(\pm[30, 60]\)\rm{mm}  \\
            R. & \(\pm[25, 45]\)\(^{\circ}\) \\
            Sc. & \([90, 110]\)\%  \\
            Sh. & \([90, 110]\)\%  \\
            \hline
            \end{tabular}
        \label{tab:ranges}
     \end{subtable}

    
     \label{tab:results}
\end{table}

\noindent\textbf{Results.} Table~\ref{tab:results} shows that, for smaller ranges, MAD is superior to the state-of-the-art learned CL measure and on par with the conventional metrics for all datasets. This is unsurprising given that NGF and NMI are effective at recovering small misalignments, as we observe in Fig.~\ref{fig2}. The conclusion of the comparison changes for the large misalignment case, where MAD demonstrates significantly superior performance in all cases, even when NMI and NGF are used in a multi-resolution scheme (Fig.~\ref{fig:qualitative_comp}). Furthermore, we observe that recovering the larger transformations results in larger errors for all image distance measures. This could be caused by the optimisation not converging or converging in a local minimum. This issue is particularly severe for the learned CL measure which demonstrates very large errors, potentially due to the fact that the classification categorical signal is not able to quantify how large the patch mismatch is. Qualitative results can be found in the Supplementary material.


\subsection{Experiment 3: Ablation Study}
To demonstrate the performance of the design choices, namely the random convolution modality augmentation, patch size and the number of patches sampled, we perform a series of ablation experiments on T1w-CT of R.I.R.E. dataset. The results are demonstrated in Table \ref{tab:ablation}. It can be seen that MAD with random convolution modality augmentation showed a lower registration error (3.69mm) than MAD without random convolution (5.07mm). We also found that a higher number of patches leads to better performance as expected. Finally, we demonstrate the method is rather robust to the choice of patch size, with slightly better results from using larger patch sizes.
 
%
\begin{table}[tbp]
        \caption{Ablation study that demonstrates the performance of each structural element: number of patches, patch size and the utility of data augmentation.
        }
        \centering
       \begin{tabular}{c c c | c c c | c c}
        \hline
        \multicolumn{3}{c|}{\textbf{\# Patches}} & \multicolumn{3}{c|}{\textbf{Patch Size}} & \multicolumn{2}{c}{\textbf{Rand. Conv.}}\\
        \hline
        \textbf{10} & \textbf{100} & \textbf{300} & \textbf{16} & \textbf{32} & \textbf{64} & \textbf{yes} & \textbf{no}\\
        \hline
        \(4.44(2.61)\) & \(4.22(2.68)\) & \(3.69(2.33)\) & \(4.39(3.09)\) & \(3.96(2.00)\)& \(3.69(2.33)\) & \(3.69(2.33)\) & \(5.07(2.64)\)\\
        \hline
        \end{tabular}
     \label{tab:ablation}
\end{table}
%




\section{Conclusion}
In this work, we design a learned contrast-agnostic geometry-focused learning-based dissimilarity measure for multi-modal image registration. Using the elegant concept of random convolutions as modality augmentation and synthetic geometry augmentation, we are able to address the challenge of learning an image distance measure for multi-modal registration without the need for aligned paired multi-modal data. We carefully study the loss landscape, the capture range and registration accuracy quantitatively in two multi-modal registration scenarios. Evaluation results demonstrate that the proposed framework outperforms established multi-modal dissimilarity measures, especially for large deformation estimation. In future works, we plan to plug the proposed measure in a deep learning registration framework, perform more tests on other multi-modal applications and adapt the framework for multi-modal deformable registration.

%
%
%
\bibliographystyle{splncs04}
\bibliography{mybibliography}

@article{Sotiras2013DeformableMI,
  title={Deformable Medical Image Registration: A Survey},
  author={A. Sotiras and C. Davatzikos and N. Paragios},
  journal={IEEE Transactions on Medical Imaging},
  year={2013},
  volume={32},
  pages={1153-1190}
}

@article{Xu2021RobustAG,
  title={Robust and Generalizable Visual Representation Learning via Random Convolutions},
  author={Z. Xu and D. Liu and J. Yang and M. Niethammer},
  journal={ICLR},
  year={2021},
}

@article{Doersch2015UnsupervisedVR,
  title={Unsupervised Visual Representation Learning by Context Prediction},
  author={C. Doersch and A. Gupta and A. Efros},
  journal={2015 IEEE International Conference on Computer Vision (ICCV)},
  year={2015},
  pages={1422-1430}
}

@inproceedings{Taleb2021MultimodalSL,
  title={Multimodal Self-Supervised Learning for Medical Image Analysis},
  author={A. Taleb and C. Lippert and T. Klein and M. Nabi},
  booktitle={Information Processing in Medical Imaging},
  year={2021}
}


@article{Zagoruyko2015LearningTC,
  title={Learning to compare image patches via convolutional neural networks},
  author={S. Zagoruyko and N. Komodakis},
  journal={2015 IEEE Conference on Computer Vision and Pattern Recognition (CVPR)},
  year={2015},
  pages={4353-4361}
}

@inproceedings{Simonovsky2016ADM,
  title={A Deep Metric for Multimodal Registration},
  author={M. Simonovsky and B. Guti{\'e}rrez-Becker and D. Mateus and N. Navab and N. Komodakis},
  booktitle={International Conference on Medical Image Computing and Computer-Assisted Intervention},
  year={2016}
}

@article{Cao2017DeformableIR,
  title={Deformable Image Registration Based on Similarity-Steered CNN Regression},
  author={X. Cao and J. Yang and J. Zhang and D. Nie and M. Kim and Q. Wang and D. Shen},
  journal={Medical image computing and computer-assisted intervention: International Conference on Medical Image Computing and Computer-Assisted Intervention},
  year={2017},
  volume={10433},
  pages={300-308}
}

@article{Haskins2018LearningDS,
  title={Learning deep similarity metric for 3D MR–TRUS image registration},
  author={G. Haskins and J. Kruecker and U. Kruger and S. Xu and P. Pinto and B. Wood and P. Yan},
  journal={International Journal of Computer Assisted Radiology and Surgery},
  year={2018},
  volume={14},
  pages={417-425}
}

@article{Cheng2015DeepSL,
  title={Deep similarity learning for multimodal medical images},
  author={X. Cheng and L. Zhang and Y. Zheng},
  journal={Computer Methods in Biomechanics and Biomedical Engineering: Imaging \& Visualization},
  year={2015}
}

@article{Haber2006IntensityGB,
  title={Intensity Gradient Based Registration and Fusion of Multi-modal Images},
  author={E. Haber and J. Modersitzki},
  journal={Medical image computing and computer-assisted intervention: International Conference on Medical Image Computing and Computer-Assisted Intervention},
  year={2006},
  volume={9 Pt 2},
  pages={726-33}
}

@article{Heinrich2012MINDMI,
  title={MIND: Modality independent neighbourhood descriptor for multi-modal deformable registration},
  author={M. Heinrich and M. Jenkinson and M. Bhushan and T. Matin and F. Gleeson and M. Brady and J. Schnabel},
  journal={Medical image analysis},
  year={2012},
  volume={16 7},
  pages={1423-35}
}

@article{Studholme1999AnOI,
  title={An overlap invariant entropy measure of 3D medical image alignment},
  author={C. Studholme and D. Hill and D. Hawkes},
  journal={Pattern Recognition},
  year={1999},
  volume={32},
  pages={71-86}
}


@article{Czolbe2020DeepSimSS,
  title={DeepSim: Semantic similarity metrics for learned image registration},
  author={Steffen Czolbe and Oswin Krause and Aasa Feragen},
  journal={ArXiv},
  year={2020},
  volume={abs/2011.05735}
}

@article{Lee2009LearningSM,
  title={Learning similarity measure for multi-modal 3D image registration},
  author={D. Lee and M. Hofmann and F. Steinke and Y. Altun and N. Cahill and B. Sch{\"o}lkopf},
  journal={2009 IEEE Conference on Computer Vision and Pattern Recognition},
  year={2009},
  pages={186-193}
}


@article{Wells1996MultimodalVR,
  title={Multi-modal volume registration by maximization of mutual information},
  author={W. Wells and P. Viola and H. Atsumi and S. Nakajima and R. Kikinis},
  journal={Medical image analysis},
  year={1996},
  volume={1 1},
  pages={35-51}
}

@article{Loeckx2010NonrigidIR,
  title={Nonrigid Image Registration Using Conditional Mutual Information},
  author={D. Loeckx and P. Slagmolen and F. Maes and D. Vandermeulen and P. Suetens},
  journal={IEEE Transactions on Medical Imaging},
  year={2010},
  volume={29},
  pages={19-29}
}

@article{Woo2015MultimodalRV,
  title={Multimodal Registration via Mutual Information Incorporating Geometric and Spatial Context},
  author={J. Woo and M. Stone and J. Prince},
  journal={IEEE Transactions on Image Processing},
  year={2015},
  volume={24},
  pages={757-769}
}

@article{Wachinger2012EntropyAL,
  title={Entropy and Laplacian images: Structural representations for multi-modal registration},
  author={C. Wachinger and N. Navab},
  journal={Medical image analysis},
  year={2012},
  volume={16 1},
  pages={1-17}
}

@article{Arar2020UnsupervisedMI,
  title={Unsupervised Multi-Modal Image Registration via Geometry Preserving Image-to-Image Translation},
  author={M. Arar and Y. Ginger and D. Danon and I. Leizerson and A. Bermano and D. Cohen-Or},
  journal={2020 IEEE/CVF Conference on Computer Vision and Pattern Recognition (CVPR)},
  year={2020},
  pages={13407-13416}
}

@article{Mok2022AffineMI,
  title={Affine Medical Image Registration with Coarse-to-Fine Vision Transformer},
  author={T. Mok and A. Chung},
  journal={2022 IEEE/CVF Conference on Computer Vision and Pattern Recognition (CVPR)},
  year={2022},
  pages={20803-20812}
}


@article{chen2019med3d,
    title={Med3D: Transfer Learning for 3D Medical Image Analysis},
    author={C. Sihong and M. Kai and Z. Yefeng},
    journal={arXiv preprint arXiv:1904.00625},
    year={2019}
}

@article{Sandkhler2018AirLabAI,
  title={AirLab: Autograd Image Registration Laboratory},
  author={R. Sandk{\"u}hler and C. Jud and S. Andermatt and P. Cattin},
  journal={ArXiv},
  year={2018},
}


@article{Kingma2015AdamAM,
  title={Adam: A Method for Stochastic Optimization},
  author={D. Kingma and J. Ba},
  journal={CoRR},
  year={2015},
}

@article{Shafto2014TheCC,
  title={The Cambridge Centre for Ageing and Neuroscience (Cam-CAN) study protocol: a cross-sectional, lifespan, multidisciplinary examination of healthy cognitive ageing},
  author={M. Shafto and L. Tyler and M. Dixon and Jason R. Taylor and J. Rowe and R. Cusack and A. Calder and W. Marslen-Wilson and J. Duncan and T. Dalgleish and R. Henson and C. Brayne and F. Matthews},
  journal={BMC Neurology},
  year={2014},
  volume={14}
}

@article{Taylor2017TheCC,
  title={The Cambridge Centre for Ageing and Neuroscience (Cam-CAN) data repository: Structural and functional MRI, MEG, and cognitive data from a cross-sectional adult lifespan sample},
  author={J. Taylor and N. Williams and R. Cusack and T. Auer and M. Shafto and M. Dixon and L. Tyler and Cam-CAN-Group and R. Henson},
  journal={Neuroimage},
  year={2017},
  volume={144},
  pages={262 - 269}
}

@article{Iglesias2011Robex,
    author = {J. Iglesias  and C. Liu  and P. Thompson and Z. Tu},
    year = {2011},
    month = {11},
    pages = {1617-1634},
    title = {Robust Brain Extraction Across Datasets and Comparison with Publicly Available Methods},
    volume = {30(9)},
    journal = {IEEE Transactions on Medical Imaging}
}

@article{Lowekamp2013TheDO,
  title={The Design of SimpleITK},
  author={B. Lowekamp and D. Chen and L. Ib{\'a}{\~n}ez and D. Blezek},
  journal={Frontiers in Neuroinformatics},
  year={2013},
  volume={7}
}

@article{Marcus2007OpenAS,
  title={Open Access Series of Imaging Studies (OASIS): Cross-sectional MRI Data in Young, Middle Aged, Nondemented, and Demented Older Adults},
  author={D. Marcus and T. Wang and J. Parker and J. Csernansky and J. Morris and R. Buckner},
  journal={Journal of Cognitive Neuroscience},
  year={2007},
  volume={19},
  pages={1498-1507}
}

@article{Hering2022Learn2RegCM,
  title={Learn2Reg: comprehensive multi-task medical image registration challenge, dataset and evaluation in the era of deep learning},
  author={A. Hering and L. Hansen and T. Mok and A. Chung and H. Siebert and S. Hager and A. Lange and S. Kuckertz and S. Heldmann and W. Shao and S. Vesal and M. Rusu and G. Sonn and T. Estienne and M. Vakalopoulou and L. Han and Y. Huang and M. Brudfors and Y. Balbastre and S. Joutard and M. Modat and G. Lifshitz and D. Raviv and J. Lv and Q. Li and V. Jaouen and D. Visvikis and C. Fourcade and M. Rubeaux and W. Pan and Z. Xu and B. Jian and F. De Benetti and M. Wodzinski and N. Gunnarsson and H. Qiu and Z. Li and C. Grossbrohmer and A. Hoopes and I. Reinertsen and Y. Xiao and B. Landman and Y. Huo and K. Murphy and B. van Ginneken and A. Dalca and M. Heinrich},
  journal={IEEE transactions on medical imaging},
  year={2022},
  volume={PP}
}

@article{Franceschi2018BilevelPF,
  title={Bilevel Programming for Hyperparameter Optimization and Meta-Learning},
  author={L. Franceschi and P. Frasconi and S. Salzo and R. Grazzi and M. Pontil},
  journal={ArXiv},
  year={2018},
}
@inproceedings{West1996ComparisonAE,
  title={Comparison and evaluation of retrospective intermodality image registration techniques.},
  author={J. West and J. Fitzpatrick and M. Wang and B. Dawant and C. Maurer and R. Kessler and R. Maciunas and C. Barillot and D. Lemoine and Andr{\'e} M. Collignon and F. Maes and P. Suetens and D. Vandermeulen and P. A. van den E. and P. Hemler and S. Napel and T. Sumanaweera and B. Harkness and D. Hill and C. Studholme and G. Malandain and X. Pennec and M. Noz and G. Maguire and M. Pollack and C. Pelizzari and R. Robb and D. Hanson and R. Woods},
  booktitle={Medical Imaging},
  year={1996}
}

@article{Muschelli2015ValidatedAB,
  title={Validated automatic brain extraction of head CT images},
  author={J. Muschelli and N. Ullman and W. Mould and P. Vespa and D. Hanley and C. Crainiceanu},
  journal={NeuroImage},
  year={2015},
  volume={114},
  pages={379-385}
}


@article{Vos2019ADL,
  title={A deep learning framework for unsupervised affine and deformable image registration},
  author={B. De Vos and F. Berendsen and M. Viergever and H. Sokooti and M. Staring and I. Isgum},
  journal={Medical Image Analysis},
  year={2019},
  volume={52},
  pages={128–143}
}

@book{Efron2016ComputerAS,
author = {E.Bradley and H. Trevor},
title = {Computer Age Statistical Inference: Algorithms, Evidence, and Data Science},
year = {2016},
publisher = {Cambridge University Press},
address = {USA},
}

@article{Cohen1969StatisticalPA,
  title={Statistical Power Analysis for the Behavioral Sciences},
  author={J. Cohen},
  journal={The SAGE Encyclopedia of Research Design},
  year={1969}
}


@article{Hering2021Learn2RegCM,
  title={Learn2Reg: Comprehensive Multi-Task Medical Image Registration Challenge, Dataset and Evaluation in the Era of Deep Learning},
  author={A. Hering and L. Hansen and T. Mok and A. Chung and H. Siebert and S. Hager and A. Lange and S. Kuckertz and S. Heldmann and W. Shao and S. Vesal and M. Rusu and G. Sonn and T. Estienne and M. Vakalopoulou and L. Han and Y. Huang and M. Brudfors and Y. Balbastre and S. Joutard and M. Modat and G. Lifshitz and Daniel R. and J. Lv and Q. Li and Vincent J. and D. Visvikis and C. Fourcade and M. Rubeaux and W. Pan and Z. Xu and B. Jian and F. De Benetti and M. Wodzinski and N. Gunnarsson and H. Qiu and Zeju Li and C. Grossbrohmer and A. Hoopes and I. Reinertsen and Y. Xiao and B. A. Landman and Y. Huo and K. Murphy and B. van Ginneken and A. Dalca and M. Heinrich},
  journal={IEEE Transactions on Medical Imaging},
  year={2021},
  volume={42},
  pages={697-712}
}
@inproceedings{he2016deep,
  title={Deep residual learning for image recognition},
  author={He, K. and Zhang, X. and Ren, S. and Sun, J.},
  booktitle={Proceedings of the IEEE conference on computer vision and pattern recognition},
  pages={770--778},
  year={2016}
}

@article{Pielawski2020CoMIRCM,
  title={CoMIR: Contrastive Multimodal Image Representation for Registration},
  author={Nicolas Pielawski and Elisabeth Wetzer and Johan Ofverstedt and Jiahao Lu and Carolina Wahlby and Joakim Lindblad and Natavsa Sladoje},
  journal={ArXiv},
  year={2020},
  volume={abs/2006.06325}
}

@inproceedings{Dey2022ContraRegCL,
  title={ContraReg: Contrastive Learning of Multi-modality Unsupervised Deformable Image Registration},
  author={Neel Dey and Jo Schlemper and Seyed Sadegh Mohseni Salehi and Bo Zhou and Guido Gerig and Michal Sofka},
  booktitle={International Conference on Medical Image Computing and Computer-Assisted Intervention},
  year={2022}
}

@inproceedings{Qin2019UnsupervisedDR,
  title={Unsupervised Deformable Registration for Multi-Modal Images via Disentangled Representations},
  author={Chen Qin and Bibo Shi and Rui Liao and Tommaso Mansi and Daniel Rueckert and Ali Kamen},
  booktitle={Information Processing in Medical Imaging},
  year={2019}
}

@article{Hoffmann2020SynthMorphLC,
  title={SynthMorph: Learning Contrast-Invariant Registration Without Acquired Images},
  author={Malte Hoffmann and Benjamin Billot and Douglas N. Greve and Juan Eugenio Iglesias and Bruce R. Fischl and Adrian V. Dalca},
  journal={IEEE transactions on medical imaging},
  year={2020},
  volume={41},
  pages={543 - 558}
}

\begin{thebibliography}{10}
\providecommand{\url}[1]{\texttt{#1}}
\providecommand{\urlprefix}{URL }
\providecommand{\doi}[1]{https://doi.org/#1}

\bibitem{Cheng2015DeepSL}
Cheng, X., Zhang, L., Zheng, Y.: Deep similarity learning for multimodal
  medical images. Computer Methods in Biomechanics and Biomedical Engineering:
  Imaging \& Visualization  (2015)

\bibitem{Cohen1969StatisticalPA}
Cohen, J.: Statistical power analysis for the behavioral sciences. The SAGE
  Encyclopedia of Research Design  (1969)

\bibitem{Czolbe2020DeepSimSS}
Czolbe, S., Krause, O., Feragen, A.: Deepsim: Semantic similarity metrics for
  learned image registration. ArXiv  \textbf{abs/2011.05735} (2020)

\bibitem{Dey2022ContraRegCL}
Dey, N., Schlemper, J., Salehi, S.S.M., Zhou, B., Gerig, G., Sofka, M.:
  Contrareg: Contrastive learning of multi-modality unsupervised deformable
  image registration. In: International Conference on Medical Image Computing
  and Computer-Assisted Intervention (2022)

\bibitem{Efron2016ComputerAS}
E.Bradley, Trevor, H.: Computer Age Statistical Inference: Algorithms,
  Evidence, and Data Science. Cambridge University Press, USA (2016)

\bibitem{Haber2006IntensityGB}
Haber, E., Modersitzki, J.: Intensity gradient based registration and fusion of
  multi-modal images. Medical image computing and computer-assisted
  intervention: International Conference on Medical Image Computing and
  Computer-Assisted Intervention  \textbf{9 Pt 2},  726--33 (2006)

\bibitem{Haskins2018LearningDS}
Haskins, G., Kruecker, J., Kruger, U., Xu, S., Pinto, P., Wood, B., Yan, P.:
  Learning deep similarity metric for 3d mr–trus image registration.
  International Journal of Computer Assisted Radiology and Surgery
  \textbf{14},  417--425 (2018)

\bibitem{he2016deep}
He, K., Zhang, X., Ren, S., Sun, J.: Deep residual learning for image
  recognition. In: Proceedings of the IEEE conference on computer vision and
  pattern recognition. pp. 770--778 (2016)

\bibitem{Heinrich2012MINDMI}
Heinrich, M., Jenkinson, M., Bhushan, M., Matin, T., Gleeson, F., Brady, M.,
  Schnabel, J.: Mind: Modality independent neighbourhood descriptor for
  multi-modal deformable registration. Medical image analysis  \textbf{16 7},
  1423--35 (2012)

\bibitem{Hering2021Learn2RegCM}
Hering, A., Hansen, L., Mok, T., Chung, A., Siebert, H., Hager, S., Lange, A.,
  Kuckertz, S., Heldmann, S., Shao, W., Vesal, S., Rusu, M., Sonn, G.,
  Estienne, T., Vakalopoulou, M., Han, L., Huang, Y., Brudfors, M., Balbastre,
  Y., Joutard, S., Modat, M., Lifshitz, G., R., D., Lv, J., Li, Q., J., V.,
  Visvikis, D., Fourcade, C., Rubeaux, M., Pan, W., Xu, Z., Jian, B., Benetti,
  F.D., Wodzinski, M., Gunnarsson, N., Qiu, H., Li, Z., Grossbrohmer, C.,
  Hoopes, A., Reinertsen, I., Xiao, Y., Landman, B.A., Huo, Y., Murphy, K., van
  Ginneken, B., Dalca, A., Heinrich, M.: Learn2reg: Comprehensive multi-task
  medical image registration challenge, dataset and evaluation in the era of
  deep learning. IEEE Transactions on Medical Imaging  \textbf{42},  697--712
  (2021)

\bibitem{Hoffmann2020SynthMorphLC}
Hoffmann, M., Billot, B., Greve, D.N., Iglesias, J.E., Fischl, B.R., Dalca,
  A.V.: Synthmorph: Learning contrast-invariant registration without acquired
  images. IEEE transactions on medical imaging  \textbf{41},  543 -- 558 (2020)

\bibitem{Iglesias2011Robex}
Iglesias, J., Liu, C., Thompson, P., Tu, Z.: Robust brain extraction across
  datasets and comparison with publicly available methods. IEEE Transactions on
  Medical Imaging  \textbf{30(9)},  1617--1634 (11 2011)

\bibitem{Lee2009LearningSM}
Lee, D., Hofmann, M., Steinke, F., Altun, Y., Cahill, N., Sch{\"o}lkopf, B.:
  Learning similarity measure for multi-modal 3d image registration. 2009 IEEE
  Conference on Computer Vision and Pattern Recognition pp. 186--193 (2009)

\bibitem{Loeckx2010NonrigidIR}
Loeckx, D., Slagmolen, P., Maes, F., Vandermeulen, D., Suetens, P.: Nonrigid
  image registration using conditional mutual information. IEEE Transactions on
  Medical Imaging  \textbf{29},  19--29 (2010)

\bibitem{Lowekamp2013TheDO}
Lowekamp, B., Chen, D., Ib{\'a}{\~n}ez, L., Blezek, D.: The design of
  simpleitk. Frontiers in Neuroinformatics  \textbf{7} (2013)

\bibitem{Mok2022AffineMI}
Mok, T., Chung, A.: Affine medical image registration with coarse-to-fine
  vision transformer. 2022 IEEE/CVF Conference on Computer Vision and Pattern
  Recognition (CVPR) pp. 20803--20812 (2022)

\bibitem{Muschelli2015ValidatedAB}
Muschelli, J., Ullman, N., Mould, W., Vespa, P., Hanley, D., Crainiceanu, C.:
  Validated automatic brain extraction of head ct images. NeuroImage
  \textbf{114},  379--385 (2015)

\bibitem{Pielawski2020CoMIRCM}
Pielawski, N., Wetzer, E., Ofverstedt, J., Lu, J., Wahlby, C., Lindblad, J.,
  Sladoje, N.: Comir: Contrastive multimodal image representation for
  registration. ArXiv  \textbf{abs/2006.06325} (2020)

\bibitem{Sandkhler2018AirLabAI}
Sandk{\"u}hler, R., Jud, C., Andermatt, S., Cattin, P.: Airlab: Autograd image
  registration laboratory. ArXiv  (2018)

\bibitem{Shafto2014TheCC}
Shafto, M., Tyler, L., Dixon, M., Taylor, J.R., Rowe, J., Cusack, R., Calder,
  A., Marslen-Wilson, W., Duncan, J., Dalgleish, T., Henson, R., Brayne, C.,
  Matthews, F.: The cambridge centre for ageing and neuroscience (cam-can)
  study protocol: a cross-sectional, lifespan, multidisciplinary examination of
  healthy cognitive ageing. BMC Neurology  \textbf{14} (2014)

\bibitem{Simonovsky2016ADM}
Simonovsky, M., Guti{\'e}rrez-Becker, B., Mateus, D., Navab, N., Komodakis, N.:
  A deep metric for multimodal registration. In: International Conference on
  Medical Image Computing and Computer-Assisted Intervention (2016)

\bibitem{Sotiras2013DeformableMI}
Sotiras, A., Davatzikos, C., Paragios, N.: Deformable medical image
  registration: A survey. IEEE Transactions on Medical Imaging  \textbf{32},
  1153--1190 (2013)

\bibitem{Studholme1999AnOI}
Studholme, C., Hill, D., Hawkes, D.: An overlap invariant entropy measure of 3d
  medical image alignment. Pattern Recognition  \textbf{32},  71--86 (1999)

\bibitem{Taylor2017TheCC}
Taylor, J., Williams, N., Cusack, R., Auer, T., Shafto, M., Dixon, M., Tyler,
  L., Cam-CAN-Group, Henson, R.: The cambridge centre for ageing and
  neuroscience (cam-can) data repository: Structural and functional mri, meg,
  and cognitive data from a cross-sectional adult lifespan sample. Neuroimage
  \textbf{144},  262 -- 269 (2017)

\bibitem{Wachinger2012EntropyAL}
Wachinger, C., Navab, N.: Entropy and laplacian images: Structural
  representations for multi-modal registration. Medical image analysis
  \textbf{16 1},  1--17 (2012)

\bibitem{Wells1996MultimodalVR}
Wells, W., Viola, P., Atsumi, H., Nakajima, S., Kikinis, R.: Multi-modal volume
  registration by maximization of mutual information. Medical image analysis
  \textbf{1 1},  35--51 (1996)

\bibitem{West1996ComparisonAE}
West, J., Fitzpatrick, J., Wang, M., Dawant, B., Maurer, C., Kessler, R.,
  Maciunas, R., Barillot, C., Lemoine, D., Collignon, A.M., Maes, F., Suetens,
  P., Vandermeulen, D., van~den E., P.A., Hemler, P., Napel, S., Sumanaweera,
  T., Harkness, B., Hill, D., Studholme, C., Malandain, G., Pennec, X., Noz,
  M., Maguire, G., Pollack, M., Pelizzari, C., Robb, R., Hanson, D., Woods, R.:
  Comparison and evaluation of retrospective intermodality image registration
  techniques. In: Medical Imaging (1996)

\bibitem{Woo2015MultimodalRV}
Woo, J., Stone, M., Prince, J.: Multimodal registration via mutual information
  incorporating geometric and spatial context. IEEE Transactions on Image
  Processing  \textbf{24},  757--769 (2015)

\bibitem{Xu2021RobustAG}
Xu, Z., Liu, D., Yang, J., Niethammer, M.: Robust and generalizable visual
  representation learning via random convolutions. ICLR  (2021)

\end{thebibliography}

\end{document}